# A new axiomatization for likelihood gambles *

**Phan H. Giang**
Computer-Aided Diagnosis & Therapy Group (CAD)
Siemens Medical Solutions Inc.
51 Valley Stream Pkwy, E51, Malvern, PA 19355 USA
*phan.giang@siemens.com*

## Abstract

This paper studies a new and more general axiomatization than one presented in [6] for preference on likelihood gambles. Likelihood gambles describe actions in a situation where a decision maker knows multiple probabilistic models and a random sample generated from one of those models but does not know prior probability of models. This new axiom system is inspired by Jensen's axiomatization of probabilistic gambles. Our approach provides a new perspective to the role of data in decision making under ambiguity.

## 1 Likelihood gambles

Likelihood gambles introduced in [5, 6] describe actions in situation of model ambiguity characterized by (1) there are multiple probabilistic models; (2) there is data providing likelihoods for the models and (3) there is no prior probability about the models.

Formally, we consider a general problem described by a tuple $(X, Y, \Theta, \mathcal{A}, \mathbf{x})$. $X, Y$ are variables describing a phenomenon of interest. $X$ is *experiment variable* whose values can be observed through experiments or data gathering (e.g. lab test results, clinical observations). $Y$ is *utility variable* whose values determine the utility of actions (e.g. stages of disease, relative size of the tumor). $\Theta$ is the set of *models* that encode the knowledge about the phenomenon. To be precise, $\Theta$ is a set of indices and knowledge is encoded in probability functions $\Pr_\theta(X, Y)$ for $\theta \in \Theta$. $\mathcal{A}$ is the set of alternative actions (e.g. surgery, radiation therapy, chemotherapy) that are functions from utility variable $Y$ to the unit interval $[0, 1]$ representing utility[1]. Finally, *evidence/data/observation* gathered on experiment variable is $X = \mathbf{x}$. A fundamental question to be answered is which among the alternative actions is the best choice given the information.

We introduce the concept of *likelihood gambles* and derive a pricing formula that will allow their comparison. Note that *given* a model $\theta \in \Theta$ and observation $\mathbf{x}$, distribution on utility variable $Y$ is $\Pr_\theta(y|\mathbf{x})$. According to the classical Bayesian decision theory actions $a \in \mathcal{A}$ are values by their expected utility

$$\mathbf{E}_\theta(a) = \int_y a(y) \Pr_\theta(y|\mathbf{x}) dy \qquad (1)$$

Beside effecting individual models through conditionalization, data also carries information about models. We will use the term *ambiguity* to denote the uncertainty about models and reserve the term *risk* for the uncertainty about variable realization. This terminology is apt because if to understand a variable is to know its probability distribution, the presence of multiple models of a variable means there are more than one ways to understand it. To this end we'll invoke a fundamental principle of statistics

**Likelihood principle (LP)** *All information about models in a data is contained in likelihood function. Moreover, two likelihood functions are equivalent if one is a scalar multiple of the other.*

The likelihood of a model is the probability that the model predicts the data: $\ell_\theta = \Pr_\theta(\mathbf{x})$. This controversial principle lies in the heart of Bayesian approach to statistics (see for example [1]). Although LP does not agree with methodology of frequentist statistics which takes into analysis not only data itself but also data gathering protocol (experiment design),

---

*I thank Bharat Rao for encouragement and support and UAI-2006 referees for their constructive comments.

[1]One should note that the use of utility rather than money to measure the consequences of actions makes the discussion simpler because the risk attitude factor has been assumed away. From now on we can think of dollar sign ($) as a symbol of the utility currency in which actions are measured.

likelihood is essential in both estimation theory and hypothesis testing theory.

A consequence of the likelihood principle is *evidence exchangeability*. For a set of models $\Theta$, two pieces of evidence $\mathbf{x}$ and $\mathbf{x}'$ are considered equivalent if $\Pr_\theta(\mathbf{x}) = c\Pr_\theta(\mathbf{x}')$ for $\theta \in \Theta$ and some constant $c$. This postulate is directly followed from the second part of likelihood principle. For example a four-face die is known to have distribution according to either one of two probability functions

|  | red | blue | green | yellow |
|---|---|---|---|---|
| $\Pr_{\theta_1}$ | 0.2 | 0.1 | 0.3 | 0.4 |
| $\Pr_{\theta_2}$ | 0.3 | 0.15 | 0.3 | 0.25 |

Evidence exchangeability means that observing "red" is equivalent to observing "blue" in terms of information relevant to question which is the true distribution of the die.

In (1), $\mathbf{E}_\theta(a)$ was interpreted as the value of an action given a model, now we turn other way around by fixing action $a$ and view $\mathbf{E}_\theta(a)$ as the value of model $\theta$ given action $a$. This switch of views is crucial to understand this work and it mirrors model-evidence switch that interprets the same quantity $\Pr_\theta(\mathbf{x})$ either as probability of evidence or as likelihood of a model.

**Assumption 1** *The value of an action in multi-model situation is a function of two arguments: its values given individual models and model ambiguity.*

Another consequence of likelihood principle is *model exchangeability*. For evidences $\mathbf{x}, \mathbf{x}'$ and two sets models $\Theta$ and $\Theta'$, an action $a$ is as valuable given $\Theta$ as action $a'$ given $\Theta'$ if there is a one-to-one map $m : \Theta \to \Theta'$ and a constant $c$ such that $\mathbf{E}_\theta(a) = \mathbf{E}_{m(\theta)}(a')$ and $\Pr_\theta(\mathbf{x}) = c\Pr_{m(\theta)}(\mathbf{x}')$ for $\theta \in \Theta$. Condition $\mathbf{E}_\theta(a) = \mathbf{E}_{m(\theta)}(a')$ guarantees that values of actions are the same under both sets of models. Now, by likelihood principle $\Pr_\theta(\mathbf{x}) = c\Pr_{m(\theta)}(\mathbf{x}')$ means that information bearing on models is the same under $\Theta$ and $\Theta'$. Since both arguments of action's value are the same, the values of the actions under $\Theta$ and $\Theta'$ must be the same.

Consider the following example. There are two experiments. In the first, two coins $\theta_1$ and $\theta_2$ are used. In the second, $\theta_1'$ and $\theta_2'$ are used.

| Coins | **head** | tail | $\mathbf{E}(a)$ |
|---|---|---|---|
| $\theta_1$ | **0.5** | 0.5 | 0.5 |
| $\theta_2$ | **0.4** | 0.6 | 0.4 |
| $a$ | 1 | 0 | |

| Coins | head | **tail** | $\mathbf{E}(a')$ |
|---|---|---|---|
| $\theta_1'$ | 0 | **1.0** | 0.5 |
| $\theta_2'$ | 0.2 | **0.8** | 0.4 |
| $a'$ | 0 | 0.5 | |

$\theta_1$ is a fair coin, $\theta_1'$ is a double-tail coin, $\theta_2$ is a bias coin of 4-to-6 head-tail ratio and $\theta_2'$ is bias coin of 2-to-8 ratio. In each experiment a coin is picked at random and is tossed. In the first one, it produces **head**. In second experiment **tail** is observed. $a$ is a lottery that pays \$1 if the next toss of the selected coin in experiment 1 produces **head** and nothing otherwise. $a'$ pays \$0.5 if the next toss of the selected coin in experiment 2 produces **tail**. The model exchangeability implies that given all that information the price one pays for $a$ should equal price paid for $a'$.

We call a pair of number $(\Pr_\theta(\mathbf{x})/\mathbf{E}_\theta(a))$ a *likelihood prospect* (here "/" is a separation symbol and not a division operation). The first number is the likelihood of the model and the second number is its value. A likelihood prospect describes a model in context of an action and a piece of data. Thus, an action in a multiple model situation is described by a collection of prospects $\{(\Pr_\theta(\mathbf{x})/\mathbf{E}_\theta(a))|\theta \in \Theta\}$.

Because of evidence exchangeabiliy we can abstract away from direct reference to evidence. Let $\Pr_\Theta(\mathbf{x}) \stackrel{def}{=} \max_{\theta \in \Theta}\{\Pr_\theta(\mathbf{x})\}$ and $\ell_\theta = \Pr_\theta(\mathbf{x})/\Pr_\Theta(\mathbf{x})$ and thus $\max_{\theta \in \Theta}\{\ell_\theta\} = 1$. Using notation $v_\theta = \mathbf{E}_\theta(a)$ we call a set of prospects with normalized likelihood values a *likelihood gamble*

$$\{(\ell_\theta/v_\theta)|\theta \in \Theta\} \qquad (2)$$

For example, actions $a, a'$ in the above example are encoded by a likelihood gamble $\{1.0/0.5, 0.8/0.4\}$.

Note that when there is certainty about model i.e., $\Theta = \{\theta\}$ the gamble corresponding to action $a$ becomes $\{1/v_\theta\}$. The value of this gamble is obviously just $v_\theta$. We want to allow that the values for a gamble be other gambles.

**Definition 1** *The set of likelihood gambles $\mathcal{G}$ is defined recursively as follows. (i) $[0,1] \subseteq \mathcal{G}$; (ii) Let $I$ be a set of indices, $\ell_i \in [0,1]$, $\max_{i \in I} \ell_i = 1$ and $g_i \in \mathcal{G}$ for $i \in I$ then $\{(\ell_i/g_i)|i \in I\} \in \mathcal{G}$; and (iii) Nothing else belongs to $\mathcal{G}$.*

This definition makes the set of likelihood gambles a *mixture set* in style of Herstein and Milnor [8] except that the mixture rule is different.

Notation convention. All numbers in this paper are between 0 and 1 (inclusive). Likelihood values are denoted by Greek letters $\alpha, \beta, \gamma$. Likelihood functions are normalized i.e., maximum likelihood is 1. Gambles are denoted by Roman letters of both upper and lower cases $F, G, H, f, g, h$. Greek letter $\theta$ is reserved for model index. Upper case $I, J$ as well as $\Theta$ are model index sets.

**Definition 2** *The* depth *of a gamble is defined recursively as follows.* $depth(x) = 0$ *for* $x \in [0, 1]$. $depth(\{\alpha_i/g_i | 1 \leq i \leq I\}) = \max\{1 + depth(g_i) | 1 \leq i \leq I\}$.

Likelihood gambles can be given *standard* interpretation by using the thought experiment technique. Two parties, "honest" Nature and Player, participate in the following game. There is a pool of hypotheses $\Theta$ that correspond to probability functions $Pr_\theta$ on variable $X$. Nature picks a hypothesis according to a *secret* protocol and generates an observation $X = x$ which is revealed to Player. A gamble $G = \{(\ell_\theta/g_\theta) | \theta \in \Theta\}$ where $(\ell_\theta)$ is the normalized likelihood function obtained from $X = x$, is a contract that honest Nature pays Player reward $g_\theta$ if $\theta$ is the hypothesis Nature actually uses. In case $g_\theta$ is a constant gamble (a value in $[0, 1]$), "paying" literally means that Player will receives that amount of utility. In case $g_\theta$ is another gamble, "paying" means offering an opportunity to play this gamble. Honesty is necessary in this game because only Nature - the paying party - knows the truth on which rewards are based. This is not a zero-sum game in a sense that Nature does not try to minimize Player's wealth. We'll study the game from Player's point of view.

From now the thought experiment interpretation will be used throughout. We are satisfied that behind each gamble is an action in a situation described by a set of models and a piece of data.

A reasonable assumption we want to make explicit here is *independence*. That is probability functions used in different gambles are mutually independent. Suppose $\theta$ and $\theta'$ are hypotheses used in gambles $G$ and $G'$. They are probability functions on variables $X$ and $X'$ respectively. Furthermore data observed are $X = x$ and $X' = x'$. For examples $\theta$ is a hypothesis about a coin ($X$) while $\theta'$ about a die ($X'$) and data is "coin lands Head" and "die rolls 3". Thus, the totality of evidence available is what we call a *compound data* $(x, x')$. Since $X' = x'$ is irrelevant to hypothesis $\theta$, under compound data $(x, x')$, the likelihood of model $\theta$ remains the same as under $X = x$ alone. However, one can join $\theta, \theta'$ together to form a new *compound model* on $(X, X')$ denoted by $(\theta, \theta')$ then the likelihood of observing the compound data under this compound hypothesis is the product of likelihoods $Pr_{(\theta, \theta')}((x, x')) = Pr_\theta(x) Pr_{\theta'}(x')$

**Definition 3** *Let $\theta$ and $\theta'$ be independent hypotheses on variables $X, X'$ respectively. Suppose $X = x$ and $X' = x'$ are observed, the likelihood of compound model* $(\theta, \theta')$ *is* $\ell_{(\theta, \theta')}(x, x') \stackrel{def}{=} Pr_\theta(X = x) Pr_{\theta'}(X' = x')$

## 2 Axiomatization

Following von Neumann-Morgenstern's approach we study a preference relation $\succeq$ on likelihood gambles and its derivative forms: equivalence relations $f \sim g$ iff $f \succeq g$ and $g \succeq f$; $f \succ g$ iff $f \succeq g$ and $g \not\succeq f$. We postulate the following axioms.

(A1) Weak order. $\succeq$ is complete and transitive.

(A2) Archimedian axiom. If $f \succ g \succ h$ then there exist normalized likelihood vectors $(\alpha_i)$ and $(\beta_i)$ such that $\{\alpha_1/f, \alpha_2/h\} \succ g$ and $g \succ \{\beta_1/f, \beta_2/h\}$.

(A3) Independence. Suppose $f \succeq g$ and let $(\alpha_i)$ be a normalized likelihood vector then for any $h$ $\{\alpha_1/f, \alpha_2/h\} \succeq \{\alpha_1/g, \alpha_2/h\}$.

(A4) Compound gamble. Suppose $f_1 = \{\beta_{1i}/f_{1i} | 1 \leq i \leq I\}$ and $f_2 = \{\beta_{2j}/f_{2j} | 1 \leq j \leq J\}$ then $\{\alpha_1/f_1, \alpha_2/f_2\} \sim \{\alpha_1\beta_{1i}/f_{1i} | 1 \leq i \leq I\} \cup \{\alpha_2\beta_{2j}/f_{2j} | 1 \leq j \leq J\}$.

(A5) Idempotence. For any index set $I$ $\{\alpha_i/f | 1 \leq i \leq I\} \sim f$

This axiom system is based on well known Jensen's axiomatization for probabilistic gambles [9]. Basic objects in von Neumann-Morgenstern approach are lotteries which are probability distributions of finite support on $[0, 1]$ - the set of outcomes. Lotteries are denoted by roman letters $p, q, r$. $p(x)$ is the probability that $p$ pays reward $x$. Compound lotteries are constructed by substitution of rewards by lotteries. Suppose $p, q$ are lotteries. $\{\alpha/p, (1-\alpha)/q\}$ denotes a compound lottery that pays $p$ with probability $\alpha$ and $q$ with probability $1 - \alpha$.

$$\{\alpha/p, (1-\alpha)/q\} \stackrel{def}{=} \alpha p + (1-\alpha)q \quad (3)$$

In other words, $\{\alpha/p, (1 - \alpha)/q\}(x) = \alpha p(x) + (1 - \alpha)q(x)$ for any $x \in [0, 1]$. Denote by $\mathcal{L}$ the set of lotteries. The following axioms are the conditions for von Neumann-Morgenstern representation theorem of preference on probabilistic gambles.

(J1) Weak order. $\succeq_{vnm}$ is complete and transitive.

(J2) Archimedian. For $p, q, r \in \mathcal{L}$ such that $p \succ_{vnm} q \succ_{vnm} r$ then there exist $\alpha, \beta$ such that $\alpha p + (1 - \alpha)r \succ_{vnm} q$ and $q \succ_{vnm} \beta p + (1 - \beta)r$

(J3) Independence. For all $p, q, r \in \mathcal{L}$ and $\alpha \in [0, 1]$ then $p \succeq_{vnm} q$ iff $\alpha p + (1-\alpha)r \succeq_{vnm} \alpha q + (1-\alpha)r$.

The similarity between $A1 - A3$ and $J1 - J3$ is evident. One just need to substitute probability function for normalized likelihood function and recognize the difference in notation for compound probabilistic lotteries $\alpha p + (1 - \alpha)q$ and compound likelihood gambles $\{\alpha/f, \beta/g\}$. Less noticeable is the fact that $A3$ is weaker than $J3$ because $A3$ is an "if" proposition whereas $J3$ is an "iff" proposition. The crucial difference between Jensen's system and our axioms is in the ways compound lotteries (gambles) are treated. In Jensen's formulation, compound lotteries are *linear combination* of lotteries. This identity is considered so basic that it is formulated as a definition rather than as an axiom which is open for debate. However, the logic behind equational definition (3) and $A4$ is the same. It requires that probability functions in different lotteries or gambles are independent so that they can be combined through multiplication. Idempotence $A5$ is also a consequence of equation (3). By definition $(\alpha p + (1 - \alpha)p)(x) = \alpha p(x) + (1 - \alpha)p(x) = p(x)$ for $x \in [0, 1]$.

The Jensenian root of our axiom system is a powerful argument for its rationality and acceptance.

We now return to preference relation $\succeq$ over likelihood gambles. In the following we'll state and prove results leading to a representation theorem for $\succeq$. We also provide a constructive proof that the axiom system is consistent or free from contradiction. Due to space limit, we'll omit most of the proofs which can be found in full version of the paper.

**Corollary 1**
(i) $A3$ implies $f \sim g \Rightarrow \{\alpha_1/f, \alpha_2/h\} \sim \{\alpha_1/g, \alpha_2/h\}$.
(ii) $A5$ implies $h \sim \{1/h\}$ for any $h$.

**Theorem 1** *Assume* $A1 - A5$ *hold. Let* $G = \{\alpha_i/g_i | i \in \Theta\}$, *for any partition* $\Theta_1 \cup \Theta_2 = \Theta$ *and* $\Theta_1 \cap \Theta_2 = \emptyset$ *such that* $\alpha_{\Theta_1} \stackrel{def}{=} \max\{\alpha_i | i \in \Theta_1\} > 0$ *then*
$$G \sim \{\alpha_{\Theta_1}/\{\alpha_i'/g_i | i \in \Theta_1\}, G_{\Theta_2}\} \qquad (4)$$
*where* $G_{\Theta_2} \stackrel{def}{=} \{\alpha_i/g_i | i \in \Theta_2\}$ *and* $\alpha_i' = \alpha_i/\alpha_{\Theta_1}$.

According to theorem 1 any subset of prospects in a gamble can be replaced by a single prospect whose reward is a gamble constructed from the prospects to be replaced.

Since the meaning of constant gambles $x \in [0, 1]$ in $\mathcal{G}$ is utility value, we'll impose an axiom to reconcile utility numerical order and preference.

(A6) Numerical order. For $x, y \in [0, 1]$, $x \geq y$ iff $x \succeq y$.

For the rest of this paper, we assume that $A1 - A6$ hold.

**Theorem 2** $1 \succeq f \succeq 0$ *for any* $f \in \mathcal{G}$.

This theorem states that all gambles are bounded by extreme values $0, 1$.

**Definition 4** *Likelihood gambles of the form* $\{\alpha/1, \beta/0\}$ *is called* canonical *gambles.*

**Theorem 3** *For any* $x \in [0, 1]$ *there exists a normalized likelihood vector* $(\alpha_0, \alpha_1)$ *such that* $x \sim \{\alpha_0/0, \alpha_1/1\}$. *Furthermore,* $1 \sim \{0/0, 1/1\}$ *and* $0 \sim \{1/0, 0/1\}$.

This theorem states that for every value $x$ there is a canonical gamble equivalent to it.

**Theorem 4** $\{\alpha_1/1, \beta_1/0\} \succeq \{\alpha_2/1, \beta_2/0\}$ *if* $\alpha_1 \geq \alpha_2$ *and* $\beta_1 \leq \beta_2$.

Because canonical gambles have the same set of rewards i.e. $\{0, 1\}$, the comparison is reduced to the likelihoods. The two conditions on $\alpha_i, \beta_i$ can be summarized by one condition on likelihood ratio. The higher the likelihood ratio in favor of getting 1 the better.

**Theorem 5**
(i) If $\{1/1, \beta_1/0\} \sim \{1/1, \beta_2/0\}$ *and* $\beta_1 > \beta_2$ *then* $\{1/1, \beta_1/0\} \sim \{1/1, \beta_3/0\}$ *for any* $\beta_3 \leq \beta_1$.
(ii) If $\{\alpha_1/1, 1/0\} \sim \{\alpha_2/1, 1/0\}$ *and* $\alpha_1 < \alpha_2$ *then* $\{\alpha_3/1, 1/0\} \sim \{\alpha_2/1, 1/0\}$ *for any* $\alpha_3 \leq \alpha_2$.
(iii) $\{1/1, \beta/0\} \succ \{\alpha/1, 1/0\}$ *for* $0 \leq \alpha < 1, 0 \leq \beta < 1$.

This theorem shows the "stickiness" of likelihood gambles. This stickiness can occur close to 0 and 1. Intuitively, when likelihood ratio in favor (against) a hypothesis is overwhelming one could consider this hypothesis true (false).

**Theorem 6** *For any* $g \in \mathcal{G}$ *there is a normalized likelihood vector* $(\alpha, \beta)$ *such that* $g \sim \{\alpha/1, \beta/0\}$.

This theorem shows that the set of canonical gambles is rich enough to represent any likelihood gamble. This is a crucial observation. One can compare any two gambles by first reducing them to their equivalent canonical forms and then using theorem 4 to compare the canonical forms.

**Definition 5**
(i) $\mathcal{B} \stackrel{def}{=} \{\langle\alpha, \beta\rangle | 0 \leq \alpha, \beta \leq 1, \max(\alpha, \beta) = 1\}$ - *a set of 2-dimensional vectors represented by points in top and right borders of the unit square.*
(ii) *For* $\langle\alpha_1, \beta_1\rangle, \langle\alpha_2, \beta_2\rangle \in \mathcal{B}$ *and scalar* $1 \leq \gamma \leq 1$ *define operations* max *and* product *as*

$$\max(\langle\alpha_1, \beta_1\rangle, \langle\alpha_2, \beta_2\rangle) \stackrel{def}{=} \langle\max(\alpha_1, \alpha_2), \max(\beta_1, \beta_2)\rangle$$
$$\gamma\langle\alpha_1, \beta_1\rangle \stackrel{def}{=} \langle\gamma\alpha_1, \gamma\alpha_2\rangle$$

*(iii)* An order $\geq_*$ on $\mathcal{B}$ is defined as

$$\langle \alpha_1, \beta_1 \rangle \geq_* \langle \alpha_2, \beta_2 \rangle \text{ iff } \alpha_1 \geq \alpha_2 \text{ \& } \beta_1 \leq \beta_2$$

*(iv)* Equality $(=)$ and strict order $(>_*)$ are defined as follows. $\langle \alpha_1, \beta_1 \rangle = \langle \alpha_2, \beta_2 \rangle$ iff $\langle \alpha_1, \beta_1 \rangle \geq_* \langle \alpha_2, \beta_2 \rangle$ and $\langle \alpha_2, \beta_2 \rangle \geq_* \langle \alpha_1, \beta_1 \rangle$. $\langle \alpha_1, \beta_1 \rangle >_* \langle \alpha_2, \beta_2 \rangle$ iff $\langle \alpha_1, \beta_1 \rangle \geq_* \langle \alpha_2, \beta_2 \rangle$ and $\langle \alpha_2, \beta_2 \rangle \not\geq_* \langle \alpha_1, \beta_1 \rangle$

Product of scalar and vector is defined as usual while maximization of vectors is pointwise maximization. It is also easy to verify

$$\langle \alpha_1, \beta_1 \rangle = \langle \alpha_2, \beta_2 \rangle \text{ iff } \alpha_1 = \alpha_2 \text{ \& } \beta_1 = \beta_2 \quad (5)$$

$$\langle \alpha_1, \beta_1 \rangle >_* \langle \alpha_2, \beta_2 \rangle \text{ iff } \begin{cases} (\alpha_1 > \alpha_2 \text{ \& } \beta_1 \leq \beta_2) \vee \\ (\alpha_1 \geq \alpha_2 \text{ \& } \beta_1 < \beta_2) \end{cases} \quad (6)$$

**Definition 6** *A relation $\succeq$ on $\mathcal{G} \times \mathcal{G}$ is represented by relation $\geq_*$ on $\mathcal{B} \times \mathcal{B}$ if there is a function $U : \mathcal{G} \to \mathcal{B}$ such that $g_1 \succeq g_2$ iff $U(g_1) \geq_* U(g_2)$.*

**Definition 7** *For $x \in [0,1]$, $\mathcal{G}_x$ is the set of canonical gambles that are indifferent to $x$ i.e., $\mathcal{G}_x \stackrel{def}{=} \{\{\alpha/1, \beta/0\} | \max(\alpha, \beta) = 1, \alpha \geq 0, \beta \geq 0, \{\alpha/1, \beta/0\} \sim x\}$*

Note that if $0 < x < 1$ then $\mathcal{G}_x$ is a singleton because otherwise theorem 5 would imply that either $x \sim 1$ or $x \sim 0$ that would contradict A6. For $x = 0$ or $x = 1$ we designate one gamble $g_{(x)}$ in $\mathcal{G}_x$ as follows.

$$g_{(0)} \stackrel{def}{=} \{\alpha^*/1, 1/0\} \text{ and } g_{(1)} \stackrel{def}{=} \{1/1, \beta^*/0\} \quad (7)$$

where $\alpha^* \stackrel{def}{=} \max\{\alpha | \{\alpha/1, 1/0\} \in \mathcal{G}_0\}$ and $\beta^* \stackrel{def}{=} \max\{\beta | \{1/1, \beta/0\} \in \mathcal{G}_1\}$.

**Theorem 7** *Suppose function $U : \mathcal{G} \to \mathcal{B}$ is defined recursively as follows. For $x \in [0,1]$*

$$U(x) \mapsto \langle \alpha_x, \beta_x \rangle \text{ where } g_{(x)} = \{\alpha_x/1, \beta_x/0\} \quad (8)$$

$$U(\{\gamma_i/g_i | 1 \leq i \leq I\}) \mapsto \max_{1 \leq i \leq I}\{\gamma_i U(g_i)\} \quad (9)$$

*If $g_1 \sim g_2$ then $U(g_1) = U(g_2)$.*

**Lemma 1** *Assume $U$ is defined by equations 8, 9, if $U(\{\alpha/1, \beta/0\}) = \langle a, b \rangle$ then $\{\alpha/1, \beta/0\} \sim \{a/1, b/0\}$.*

**Theorem 8** *If relation $\succeq$ satisfies axioms $A1 - A6$ then it is represented by $\geq_*$.*

**Theorem 9** *For $u : \mathcal{G} \to \mathcal{B}$ that has the following properties*

*(i)* $u(x) \geq_* u(y)$ iff $1 \geq x \geq y \geq 0$
*(ii)* $u(\{\gamma_i/g_i | 1 \leq i \leq I\}) = \max_i\{\gamma_i u(g_i)\}$

*relation $\succeq_u$ defined by $g_1 \succeq_u g_2$ iff $u(g_1) \geq_* u(g_2)$ satisfies axioms $A1 - A6$.*

## 3 Ambiguity attitude

In this section, we'll study how to characterize a utility function. It is clear from eq. (9) that a utility function is completely determined by its values on constant gambles. Recall that $U(x) = \langle a, b \rangle$ implies $x \sim \{a/1, b/0\}$. To be specific let us consider two coins $\theta_1$ and $\theta_2$ with $\Pr_{\theta_1}(\text{Head}) \propto a$ and $\Pr_{\theta_2}(\text{Head}) \propto b$.[2] One of them is picked and tossed landing Head. $\{a/1, b/0\}$ is the contract that pays \$1 if the selected coin is $\theta_1$, and \$0 otherwise. For Player holding $U(x) = \langle a, b \rangle$, the *fair* price for contract $\{a/1, b/0\}$ is $x$.

Imagine for a moment that our Player is a pure Bayesianist who in order to calculate a fair price of the contract would assign prior probability $\Pr(\theta_1) = \rho$ that $\theta_1$ would be picked ($\theta_2$ would be selected with probability $1 - \rho$). With this prior information, using Bayes theorem, posterior odd of $\theta_1$ is

$$\text{odd}_{\text{post}} = \frac{\Pr(\theta_1)}{\Pr(\theta_2)} \frac{\Pr_{\theta_1}(\text{Head})}{\Pr_{\theta_2}(\text{Head})} = \frac{\rho}{1-\rho} \frac{a}{b} \quad (10)$$

The (posterior) probability that the selected coin is $\theta_i$ after observing Head is $\Pr(\theta_1|\text{Head}) = \text{odd}_{\text{post}}/(1 + \text{odd}_{\text{post}})$ and $\Pr(\theta_2|\text{Head}) = 1/(1 + \text{odd}_{\text{post}})$. Now the contract that pays \$1 if $\theta_1$ and \$0 if $\theta_2$ is a probabilistic gamble whose expected value is

$$\Pr(\theta_1|\text{Head}) * 1 + \Pr(\theta_0|\text{Head}) * 0 = \frac{\text{odd}_{\text{post}}}{1 + \text{odd}_{\text{post}}} \quad (11)$$

Since $x$ is a fair price for the contract $\{a/1, b/0\}$

$$x = \frac{\text{odd}_{\text{post}}}{1 + \text{odd}_{\text{post}}} \quad (12)$$

Solving (12) for $\text{odd}_{\text{post}}$, we have

$$\text{odd}_{\text{post}} = \frac{x}{1-x} \quad (13)$$

By eqs. 10 and 13, taking logarithm and using logit abbreviation

$$\text{logit}(z) \stackrel{def}{=} \ln\left(\frac{z}{1-z}\right) \text{ for } z \in (0,1) \quad (14)$$

we have the basic equation for canonical gambles:

$$\text{logit}(\rho) + \ln\left(\frac{a}{b}\right) = \text{logit}(x). \quad (15)$$

The significance of equation (15) is that it allows inference of *unobserved* prior from *observed* pricing of a likelihood gamble. For this reason, $\rho$ is called *implicit prior*. It is important to note that derivation of $\rho$ does not hinge on prior information about how likely each coin is selected. Due to model exchangeability property, the coins can be substituted by dice or in fact any probability devices as long as likelihood ratio obtained from data remains unchanged. Since $\rho$ is independent of physical devices actually used in a gamble, it must be an attribute of the decision maker and therefofe we use term *ambiguity attitude* for it. In particular logit($\rho$) is *ambiguity premium*. In this terminology equation (15) reads: *the utility logit equals the sum of logarithm of likelihood ratio and ambiguity premium.*

---

[2] We use $\propto$ rather than equality because $(a, b)$ is supposed to be normalized likelihood vector.

Intuitive meaning of ambiguity premium can be seen in eq. (15) by setting $a = b$. Then logit($\rho$) = logit($x$). In other words, ambiguity premium is the logit of the price that Player pays for a *fair* likelihood gamble in which the available evidence equally supports possibilities of getting $1 and getting $0. Since logit($\rho$) can be anywhere in the real line, an obvious classification is made by comparing with 0. logit($\rho$) > 0 signifies *ambiguity seeking*. Player is willing to pay more than $0.5 for the fair gamble. logit($\rho$) < 0 signifies *ambiguity averse*, the price for the fair gamble is less than $0.5. logit($\rho$) = 0 is *ambiguity neutral*, the price Player pays is $0.5.

## 4 Pricing under constant ambiguity aversion

It should be noted here that ambiguity premium is calculated on the basis of an observed pricing. In principle, that value can vary across gambles as long as the monotonicity condition $x \geq y$ iff $U(x) \geq_* U(y)$ is satisfied. In order to arrive at a closed form pricing formula we will assume ambiguity premium to be constant.

**Assumption 2** *Each decision maker has a constant ambiguity premium.*

We want to find pricing formula for likelihood gamble $G = \{(\ell_i/v_i)|1 \leq i \leq I\}$ that represents value of an action in a multiple model situation. Denote by $c$ the ambiguity attitude of Player guaranteed by assumption 2 i.e. $c = \text{logit}(\rho)$. Plug that into eq. (15) and taking into account the fact that $(a,b)$ is a normalized likelihood vector

$$\begin{cases} c + \ln\left(\frac{a}{b}\right) &= \text{logit}(x) \\ \max(a,b) &= 1 \end{cases} \quad (16)$$

Solve (16) for $a,b$ we have

$$\begin{cases} a &= \min(1, \exp(\text{logit}(x) - c)) \\ b &= \min(1, (\exp(\text{logit}(x) - c))^{-1}) \end{cases} \quad (17)$$

By eq. (9) we have

$$U(G) = \max_i(\ell_i U(v_i)) = \max_i(\ell_i \langle a_i, b_i \rangle)$$

$$= \left\langle \max_i(\ell_i a_i), \max_i(\ell_i b_i) \right\rangle \quad (18)$$

where $a_i, b_i$ are determined by equation (17) substituting $x$ by $v_i$.

Let $v$ be the price of $G$. That is $v$ is a constant gamble such that $v \sim G$. Using equation (15) again we have

$$\text{logit}(v) = \ln\left(\frac{\max_i(\ell_i a_i)}{\max_i(\ell_i b_i)}\right) + c = A \quad (19)$$

Denoting the right hand side of (19) by $A$ and solve this equation for $v$ we have *pricing formula* which has the form of the logistic function:

$$v = \frac{\exp(A)}{1 + \exp(A)} = \frac{1}{1 + \exp(-A)} \quad (20)$$

## 5 An example

Let us consider an example. There is a coin whose bias is unknown. Let $p$ be the probability of Head. The fact that bias is unknown implies $p \in [0,1]$. A gambler is allowed to observe $m$ tosses. After that she is asked to bet on $(m+1)^{\text{th}}$ toss. Specifically, she is asked to name a price for gamble $a$ that pays $1 if $(m+1)^{\text{th}}$ toss turns up Head and $0 if it turns Tail.[3] That is all she knows about the gamble.

Thus, the set of plausible models (indexed by coin bias) $\Theta = [0,1]$. $X \sim \text{B}(m,p)$ - binomial distribution of $m$ trials and probability of success $p$. $Y$ is a Bernoulli variable with probability of success $p$ i.e., $Y \sim \text{B}(1,p)$. $x$ is the observed number of successes in $m$ experimental tosses. Action $a$ is defined as $a(\text{success}) = 1$ and $a(\text{failure}) = 0$. Clearly, under model $p$ the expected utility of $a$ is $\mathbf{E}_p(a) = p$.

A Bayesian approach in this case would call for the use of noninformative prior on $p$ which can be (1) uniform distribution; (2) Jeffreys prior and reference prior which are the same in this case; and (3) prior proposed by Novick and Hall. The posterior distributions from these priors and data $X = x$ are Beta distributions. See Table 1 cited from [14]. Calculation shows that the value of $a$ under uniform prior, Jeffreys/reference prior and Novick-Hall prior are $(x+1)/(m+2)$, $(x+.5)/(m+1)$ and $x/m$ respectively.

In our approach, action $a$ is represented by gamble $G = \{\ell_p/p|p \in [0,1]\}$ where $\ell_p$ is the normalized likelihood of model $p$ from data $X = x$.

$$\ell_p \propto \binom{m}{x} p^x (1-p)^{(m-x)} \quad (21)$$

---

[3] Recall that $ is utility currency and gambler's risk attitude is assumed away.

| Method | Prior density (unnormalized) | Posterior density $\Pr(p\|x)$ | $\mathbf{E}(p\|X=x)$ |
|---|---|---|---|
| Uniform | 1 | $Be(p\|x+1, m-x+1)$ | $(x+1)/(m+2)$ |
| Jeffreys/Reference | $p^{-1/2}(1-p)^{-1/2}$ | $Be(p\|x+.5, m-x+.5)$ | $(x+.5)/(m+1)$ |
| Novick-Hall | $p^{-1}(1-p)^{-1}$ | $Be(p\|x, m-x)$ | $x/m$ |

Table 1: Noninformative priors for binomial model (Yang and Berger 1998)

| # succ | Likli-hood | Uniform prior | Jeffrey Ref. prior | Novick Hall prior |
|---|---|---|---|---|
| 0 | 0.0373 | 0.0833 | 0.0455 | 0.0000 |
| 1 | 0.1476 | 0.1667 | 0.1364 | 0.1000 |
| 2 | 0.2489 | 0.2500 | 0.2273 | 0.2000 |
| 3 | 0.3494 | 0.3333 | 0.3182 | 0.3000 |
| 4 | 0.4498 | 0.4167 | 0.4091 | 0.4000 |
| 5 | 0.5000 | 0.5000 | 0.5000 | 0.5000 |
| 6 | 0.5502 | 0.5833 | 0.5909 | 0.6000 |
| 7 | 0.6506 | 0.6667 | 0.6818 | 0.7000 |
| 8 | 0.7511 | 0.7500 | 0.7727 | 0.8000 |
| 9 | 0.8524 | 0.8333 | 0.8636 | 0.9000 |
| 10 | 0.9627 | 0.9167 | 0.9545 | 1.0000 |

Table 2: Pricing by likelihood method and Bayesian method with various priors

For $c = 0$, $\alpha_p = \min(1, \exp(\text{logit}(p)))$ and $\beta_p = \min(1, 1/\exp(\text{logit}(p)))$ Thus, $\alpha = \max_{0 \leq p \leq 1}(\ell_p \alpha_p)$ and $\beta = \max_{0 \leq p \leq 1}(\ell_p \beta_p)$. By eq. 20 $v = (\alpha/\beta)/(1 + \alpha/\beta)$.

Table 2 presents the values for the gamble found by likelihood pricing method and Bayesian method with various non-informative priors when number of trials $m = 10$.

## 6 Related works

The problem of ambiguity has been extensively studied in Economics literature. Knight [11] is often attributed the distinction between risk and uncertainty. The former corresponds to situations where a unique probability function on relevant event space is known while the later, Knightian uncertainty, corresponds to situations where there are more than one plausible probability functions. Ellsberg [4] who introduces the term ambiguity has shown that certain patterns of rational behavior are not compatible with Savage's axioms that are necessary and sufficient condition for the existence of a unique subjective probability function. Schmeidler [12] describes a decision theoretic model of ambiguity. He introduces *Choquet expected utility* (CEU) model whereby acts are evaluated by Choquet expected utility wrt a non-additive uncertainty measure. Gilboa and Schmeidler [7] suggest that a decision maker can entertain a set of (additive) probability functions. Each act is evaluated to the minimal expected utility (MEU). In an important special case, CEU and MEU are equivalent. More recently Klibanoff et al [10] propose a model in which an act $f$ is evaluated by $\mathbf{E}_\mu \phi(\mathbf{E}_\pi(u \circ f))$ where $u$ is (first order) utility function reflecting risk attitude, $\pi \in \Pi$ - the set of relevant probability functions and $\mu$ is a (second order) probability function over $\Pi$ and $\phi$ is a second order utility function reflecting ambiguity attitude. This model basically applies subjective expected utility (SEU) twice. Given probability function $\pi$ and (first order) utility function $u$, an act is evaluated to $v_{f,\pi} \stackrel{def}{=} \mathbf{E}_\pi(u \circ f)$. Given second order probability function $\mu$ (prior probability) over the set of probability functions $\Pi$ and second order utility function $\phi$, the act is evaluated to $v_f = \mathbf{E}_\mu v_{f,\pi}$.

In AI literature, the problem of uncertainty is mostly discussed in terms of representation and reasoning. Decision making process for various reasons does not get attention it deserves. Chu and Halpern [2, 3] formulate a general decision rule called *generalized expected utility* (GEU). The key construct of GEU is the notion of expectation domain where combination of uncertainty with utility and combination of utility values are made.

This paper further develops a decision theory for likelihood gambles introduced in [5, 6]. What distinguishes this approach from others is the role of data. In MEU or GEU decision models acts are evaluated wrt uncetainty represented by a set of probability functions which is assumed as given. There is no place for data in decision making phase. Any data that exists has been absorbed in the process that results in the probability functions. In our approach, uncertainty is represented by a likelihood function that is the direct result of relevant probability functions *and* data.

In our previous works likelihood information is interpreted as belief function [13]. The axioms in this paper are based on Jensenian axioms for probabilistic gambles. They are intuitively appealing and are justified by semantics of likelihood as a measure of evidence. Most of axioms used previously are theorems followed from axioms $A1 - A6$ in this paper. More important we show that this new axiom system is more general than one used before. The qualitative monotonicity

axiom in [5, 6] requires that

$$\{\alpha_1/1, \beta_1/0\} \succeq \{\alpha_2/1, \beta_2/0\} \text{ iff } \alpha_1 \geq \alpha_2 \ \& \ \beta_1 \leq \beta_2 \tag{22}$$

In theorem 5, we show that axioms $A1 - A6$ permits $\{1/1, \beta_1/0\} \sim \{1/1, 0/0\}$ with $\beta_1 > 0$ and $\{\alpha_1/1, 1/0\} \sim \{0/1, 1/0\}$ with $\alpha_1 > 0$. This preference with stickiness obviously does not satisfy qualitative monotonicity eq. (22) but is useful for expressiveness. It corresponds to model selection behavior that picks a model when likelihood ratio in its favor exceeding a pre-defined threshold.

An important feature in our approach is the separation between ambiguity attitude and risk attitude as well as separation between ambiguity representing decision maker information and ambiguity attitude representing her tastes. The model by Klibanoff et al also has this property. But in our approach, the notion of ambiguity attitude has different interpretation than one suggested in [10] which is, in fact, the second order risk attitude.

## 7 Conclusion and further work

This paper describes a novel approach to decision making under ambiguity. We provide a new and more general axiomatization of preference over likelihood gambles. We argue that likelihood gambles adequately describe actions available to decision makers in situation where she knows experimental data and the set of plausible models but not prior probability on models. The axioms adopted for likelihood gambles have the same logic and intuition as widely accepted Jensen's axioms for probabilistic gambles. We show that preference relation over likelihood gambles is represented by a utility function mapping gambles into 2-dimensional vectors. We define the notion of ambiguity aversion attitude to describe decision maker's taste. Under constant ambiguity aversion assumption, we derive pricing formula for likelihood gambles. This formula has a form of a familiar logistic function. Our approach provides a new perspective on the role of data in decision making under ambiguity. Because subjective prior probability on models is not required to find solutions using this approach, the solutions can claim to be objective and be useful as a common ground for people with different personal prior probabilities.